\documentclass[conference]{IEEEtran}
\IEEEoverridecommandlockouts
\usepackage{cite}
\usepackage{amsmath,amssymb,amsfonts}
\usepackage{algorithmic}
\usepackage{graphicx}
\usepackage{textcomp}
\usepackage{xcolor}
\usepackage[table]{xcolor} 
\usepackage{cellspace}
\usepackage{booktabs}
\usepackage{multirow} 
\usepackage{hyperref}
\usepackage[caption=false,font=footnotesize]{subfig}
\usepackage{placeins}
\usepackage{dblfloatfix}

\def\BibTeX{{\rm B\kern-.05em{\sc i\kern-.025em b}\kern-.08em
    T\kern-.1667em\lower.7ex\hbox{E}\kern-.125emX}}
\begin{document}

\title{Parameter-Efficient Subspace Decoupling ViT for Mitigating Multi-Task Negative Transfer in Histological Scoring}

\author{
\IEEEauthorblockN{
Youhan Huang\textsuperscript{1*},
Jiajun Li\textsuperscript{1*$\dagger$},
Yilin Fang\textsuperscript{1*},
Shuai Wang\textsuperscript{2},
Chuheng Li\textsuperscript{3}
}
\IEEEauthorblockA{
\textsuperscript{1}Beijing University of Posts and Telecommunications, Beijing, China \\
\textsuperscript{2}Beijing University of Chemical Technology, Beijing, China \\
\textsuperscript{3}Capital Medical University, Beijing, China
}
\thanks{* Equal contribution.}
\thanks{$\dagger$ Corresponding author. E-mail: lijiajun@bupt.edu.cn}
}

\maketitle

\begin{abstract}
Histological scoring is essential for diagnosing \textsc{N}on\textsc{a}lcoholic \textsc{F}atty \textsc{L}iver \textsc{D}isease (NAFLD), yet its automation remains challenging due to the high annotation cost and negative transfer among the strongly correlated \textsc{N}AFLD \textsc{A}ctivity \textsc{S}core (NAS) indicators in multi-task learning. To address this issue, we propose a subspace-decoupled multi-task \textsc{V}ision \textsc{T}ransformer (ViT) that integrates lightweight task-specific Adapters with orthogonality-based constraints. This design constructs independent feature subspaces for steatosis, ballooning, and inflammation, effectively reducing task interference while retaining shared representations. We further construct a curated multi-task mouse NAFLD histology dataset with expert annotations for all NAS components. Experimental results demonstrate that the proposed method improves multi-task stability and generalization with substantially reduced computational cost compared to training separate single-task models. The code and the curated dataset have been prepared and will be made publicly available upon acceptance to support reproducibility.

\end{abstract}


\section{Introduction}
\label{sec:intro}

Non-alcoholic fatty liver disease (NAFLD) is a highly prevalent chronic liver condition closely associated with insulin resistance and metabolic disorders. Its clinical management relies on long-term assessment of hepatic steatosis, inflammation, and fibrosis. However, conventional histological evaluation requires repeated manual examination of biopsy slides, which is labor-intensive and subject to substantial inter-observer variability \cite{Younossi1998NAFLDVariability}. These limitations motivate the development of computer-aided diagnosis (CAD) systems for automated and objective liver pathology assessment.

Among the pathological stages of NAFLD, Non-Alcoholic Steatohepatitis (NASH) plays a decisive role in disease severity and progression. In clinical practice, the Non-Alcoholic Steatohepatitis Activity Score (NAS) serves as the standard grading tool for distinguishing simple steatosis from true NASH by jointly evaluating steatosis, hepatocellular ballooning, and lobular inflammation \cite{kleiner2005nas}. Consequently, reliable automated NAS estimation is a critical prerequisite for scalable and practical NAFLD CAD systems.

Recent studies have explored automated NAFLD assessment using diverse data sources and learning paradigms, including CT-based prediction of steatosis and fibrosis \cite{9288092}, patch-level histopathological analysis \cite{QU2021106153}, and unsupervised or weakly supervised approaches to reduce annotation cost \cite{Karagoz2023}. In parallel, deep learning methods have improved the objectivity and reproducibility of biopsy-based grading \cite{heinemann2019deep,heinemann2022deep}. Despite this progress, many existing approaches either rely on expensive imaging modalities or treat individual NAS components independently, providing limited insight into their joint modeling.

\begin{figure}[t]
    \centering
    \vspace{-5mm}
    \subfloat[Trainable parameter count and percentage for full fine-tuning and our Adapter-based tuning on the same Swin-T backbone.]{
        \includegraphics[width=0.46\linewidth]{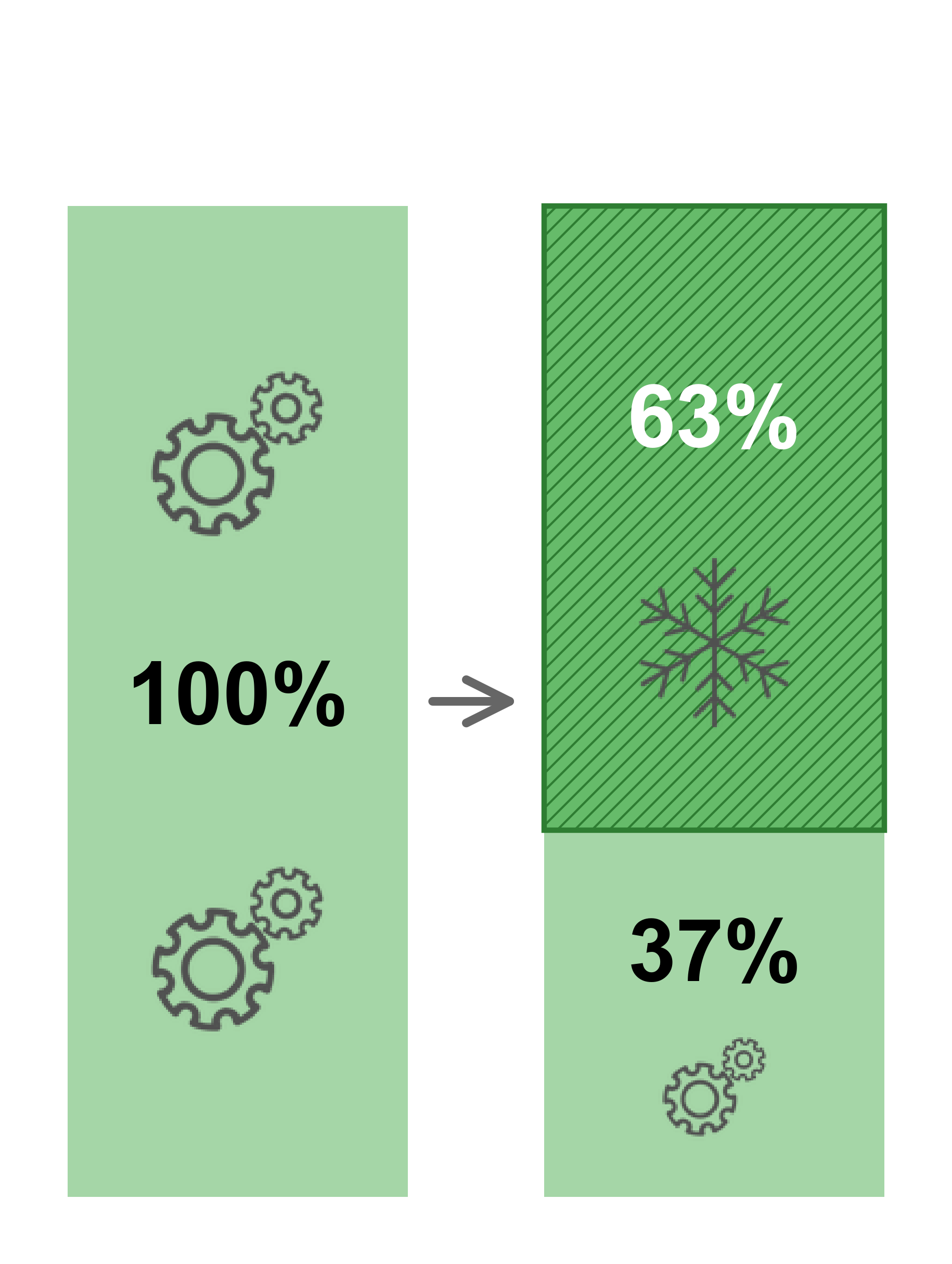}
        \label{fig:params}
    }
    \hfill
    \subfloat[Pearson correlation matrix of steatosis, ballooning, and inflammation scores on our NAFLD dataset.]
    {
        \includegraphics[width=0.46\linewidth]{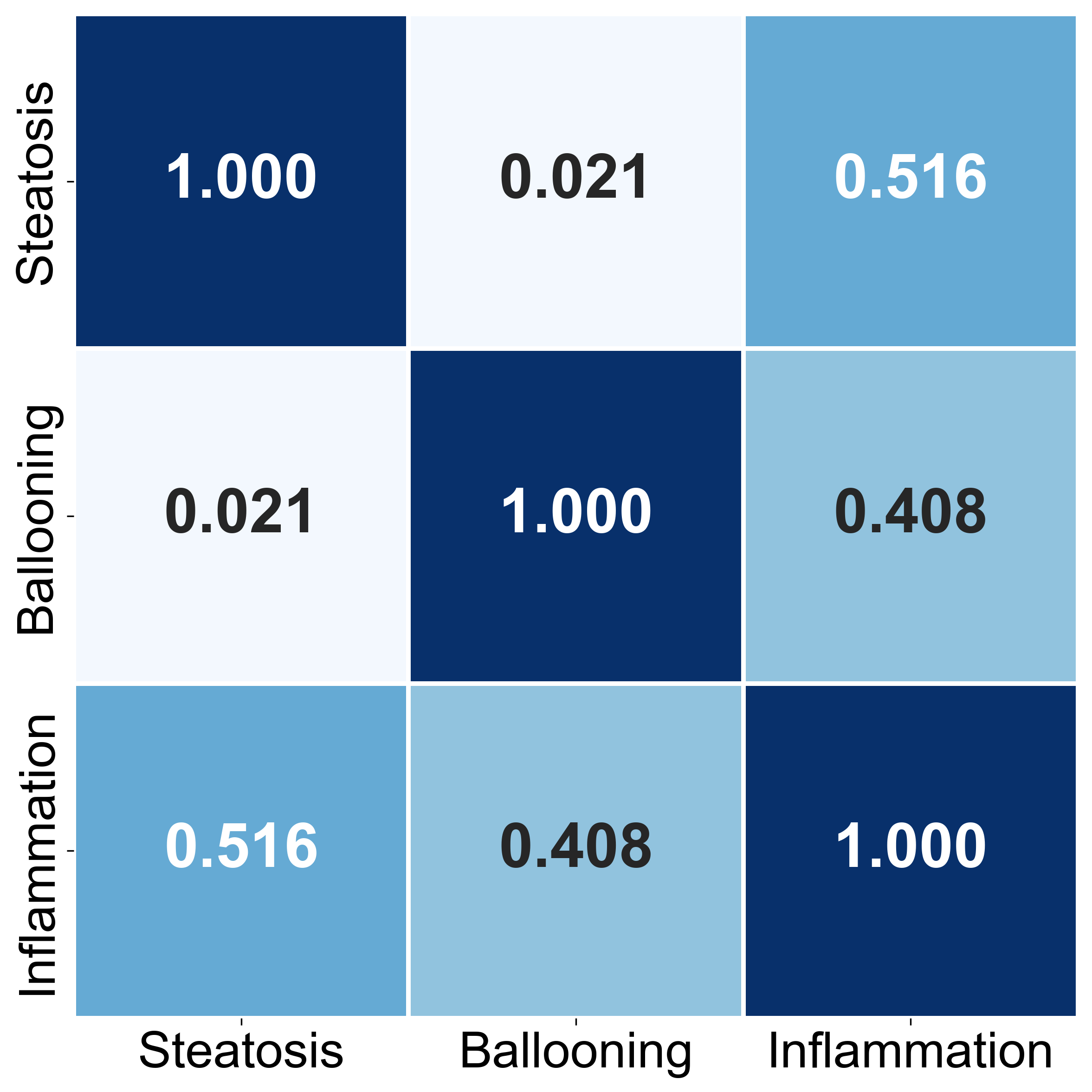}
        \label{fig:dataset_cor}
    }
    \caption{Teaser illustrating the motivation and design of the proposed parameter-efficient multi-task framework.}
    \vspace{-6mm}
    \label{fig:teaser}
\end{figure}

From a learning perspective, multi-task learning (MTL) has been widely adopted in medical image analysis and computational pathology to jointly model related objectives \cite{kamiri2025multi,liuDSCADualStreamNetwork2023,shaoMultiInstanceMultiTaskLearning2024a,grahamOneModelAll2022}. However, NAFLD histology poses a distinctive challenge. We observe that NAS components exhibit strong pairwise correlations, as consistently revealed by Pearson and Spearman correlation matrices computed on our dataset and an open-source cohort \cite{OSF_p48rd_2019} (Fig.~\ref{fig:dataset_cor}). Such correlations imply that naïvely sharing representations across tasks can bias optimization toward easier objectives, inducing task interference and negative transfer, particularly under imbalanced task difficulty and label distributions.

Recent progress in Vision Transformers (ViTs) has demonstrated strong capability in capturing long-range context and global tissue structures in histopathology and medical imaging \cite{atabansi2023survey,liu2025transformers}. These properties make ViTs well-suited for modeling heterogeneous morphological patterns in NAFLD. Nevertheless, most existing ViT-based methods focus on single-task settings and rely on full fine-tuning, offering limited support for parameter-efficient adaptation and explicit mitigation of task interference.

\begin{figure*}[!t]
    \centering
    \includegraphics[width=\linewidth]{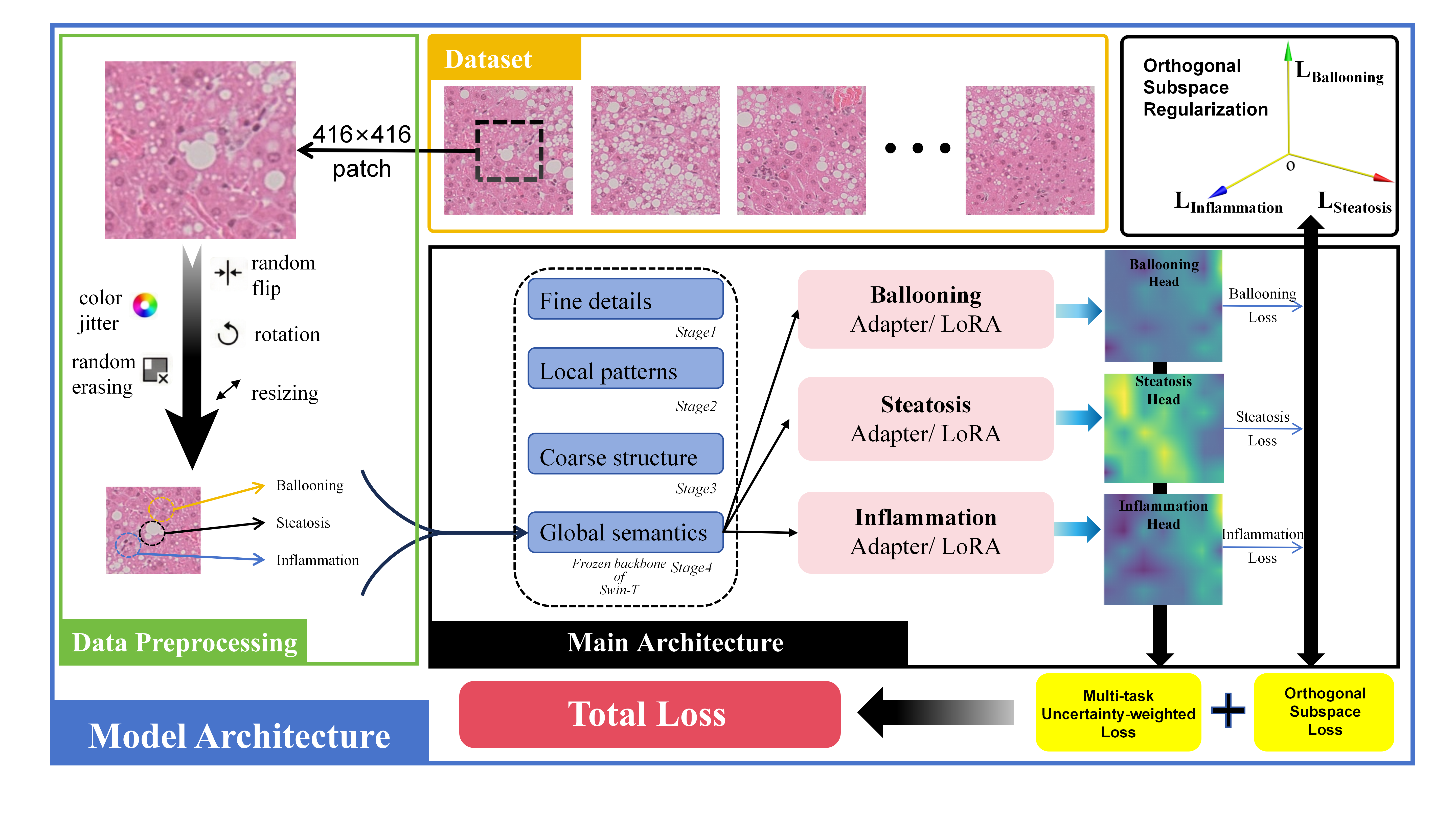}
    \vspace{-12mm}
    \caption{Overview of our parameter-efficient multi-task Swin-T framework for NAFLD histological scoring. A frozen Swin-T backbone with task-specific Adapter/LoRA branches predicts Ballooning, Steatosis, and Inflammation from augmented H\&E patches, and is trained with uncertainty-weighted multi-task loss and orthogonal subspace regularization.}
    \vspace{-7mm}
    \label{fig:architecture}
\end{figure*}

Inspired by the above observation, we propose a parameter-efficient multi-task framework based on a Swin-ViT backbone with task-specific Adapter modules for NAS scoring. By inserting lightweight Adapters into higher transformer layers, each NAS indicator is encouraged to occupy an independent high-level semantic subspace while still benefiting from shared low-level representations. To further suppress negative transfer induced by task correlations, we introduce an orthogonality-based subspace decoupling constraint that explicitly reduces correlations among task-specific representations. Extensive experiments demonstrate that the proposed framework achieves improved multi-task stability and generalization with significantly fewer parameters than full fine-tuning baselines.

Our contributions are summarized as follows:
\begin{itemize}
  \item We propose a parameter-efficient multi-task Vision Transformer framework for NAS prediction, which integrates task-specific Adapters with orthogonal subspace regularization and uncertainty-weighted optimization to mitigate negative transfer among correlated histological scores.
  
  \item Extensive experiments on an internal NAFLD dataset and a public benchmark demonstrate that the proposed framework achieves competitive or improved performance compared to single-task and full fine-tuning baselines, while requiring substantially fewer trainable parameters.
  
  \item We construct and release a curated patch-level NAFLD histology dataset with complete annotations for all NAS components, providing a benchmark for multi-task learning in liver pathology.
\end{itemize}

\section{Methodology}
\label{sec:method}

In this section, we present a parameter-efficient multi-task framework built on a shared Swin Transformer backbone. We first outline the overall architecture, then describe the task-specific Adapter modules, followed by the orthogonal decoupling loss that explicitly mitigates negative transfer. We additionally adopt an uncertainty-based weighting strategy to stabilize multi-task optimization under task imbalance.

\subsection{Overall Architecture}

Fig.~\ref{fig:architecture} shows the overall architecture of our framework, which builds on a shared Swin-T backbone pre-trained on ImageNet~\cite{Liu_2021_ICCV} with parameter-efficient fine-tuning: for each task, lightweight Adapter/LoRA modules and a task-specific head are added while the backbone stays frozen, sharing low-level features yet limiting cross-task interference.

\subsection{Task-Specific Adapter Modules}

For each NAS sub-task, we insert small bottleneck Adapter layers into the MLP branch of the last Swin-T stage, on top of the frozen ImageNet-pretrained backbone. Each Adapter takes the $d$-dimensional features from the backbone, projects them into a much lower-dimensional space ($r \ll d$), applies Layer Normalization and a GELU non-linearity, then projects back to $d$ dimensions and adds the result to the original features via a residual connection. The up-projection is zero-initialized so that, at the start of fine-tuning, each Adapter behaves as an (almost) identity mapping and does not disturb the pretrained representation; as training progresses, it gradually learns task-specific adjustments.

Each task maintains its own set of small Adapter parameters, keeping the number of trainable parameters low while allowing separate “last-mile” transformations for steatosis, ballooning, and inflammation. We use this module because full fine-tuning of Swin-T is expensive and tends to entangle tasks in a single shared space, whereas bottleneck Adapters provide a parameter-efficient way to steer the backbone along a few task-specific directions that can later be regularized by our orthogonal subspace loss.

\subsection{Orthogonal Decoupling Loss}

To explicitly control how different tasks use the shared backbone, we regularize the down-projection matrices of the task-specific Adapters to be mutually orthogonal. Let $A_t \in \mathbb{R}^{d\times r}$ denote the Adapter down-projection for task $t$, whose columns span its low-dimensional feature subspace. We define the orthogonal decoupling loss as
\begin{equation}
L_{\text{ortho}} = \sum_{i \neq j} \left\| A_i^\top A_j \right\|_F^2,
\end{equation}
where $\|\cdot\|_F$ denotes the Frobenius norm. In practice, we compute $L_{\text{ortho}}$ for all task pairs and sum it over the Transformer layers where Adapters are inserted, and control its overall strength with a weighting hyperparameter $\lambda$.

We introduce this module because NAS sub-scores (steatosis, ballooning, inflammation) are highly correlated: without constraints, easier tasks (e.g., steatosis) tend to dominate shared features and degrade harder ones (ballooning, inflammation). Geometrically, $A_i^\top A_j$ measures how much the Adapter subspace of task $i$ overlaps with that of task $j$; minimizing $\|A_i^\top A_j\|_F^2$ pushes these subspaces toward orthogonality. In this way, each Adapter is encouraged to carve out its own feature subspace, so gradients from different tasks act on largely disjoint directions, mitigating negative transfer while complementing the loss-level balancing from uncertainty weighting.

\begin{figure}[t]
    \centering
    \setlength{\abovecaptionskip}{2pt}
    \setlength{\belowcaptionskip}{-4pt}
    
    \subfloat[Steatosis]{
        \includegraphics[width=0.95\linewidth,
                         trim=0 0 0 23pt,clip]
                        {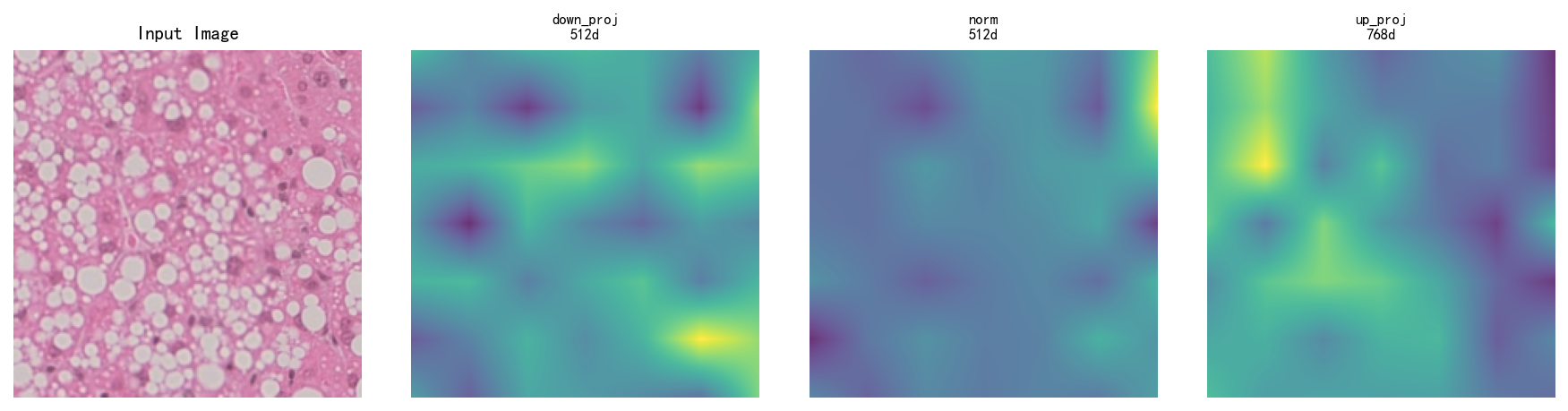}
    }\vspace{-0.4em}

    \subfloat[Ballooning]{
        \includegraphics[width=0.95\linewidth,
                         trim=0 0 0 23pt,clip]
                        {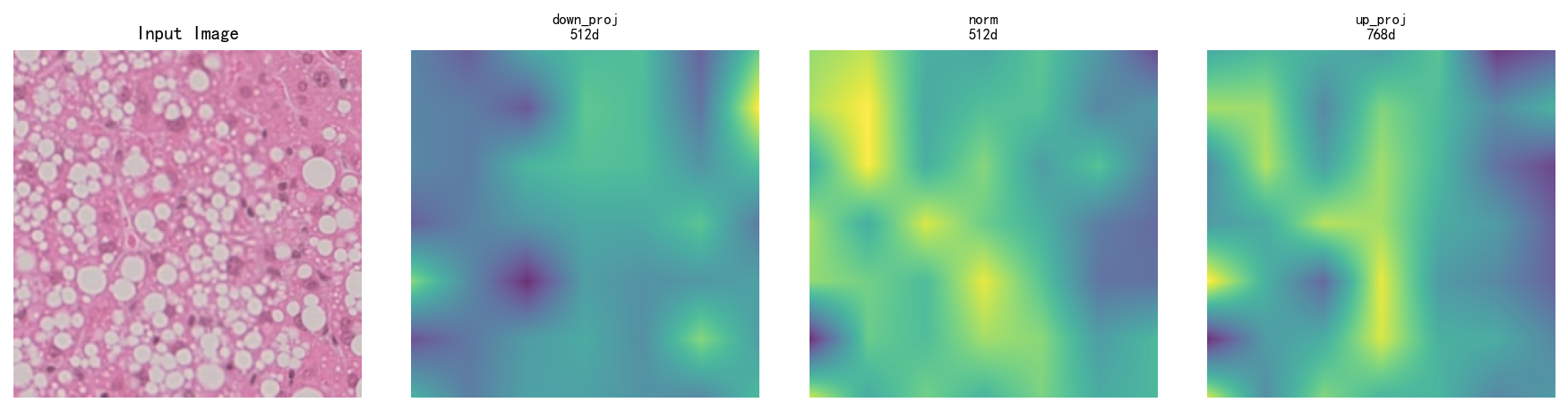}
    }\vspace{-0.4em}

    \subfloat[Inflammation]{
        \includegraphics[width=0.95\linewidth,
                         trim=0 0 0 23pt,clip]
                        {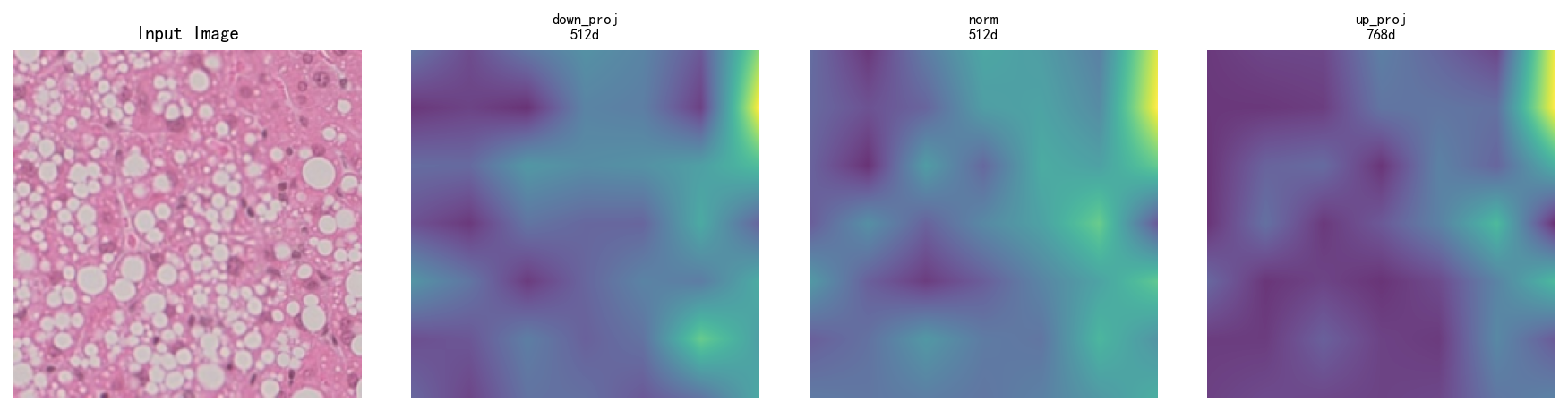}
    }\vspace{-0.2em}

    \caption{Spatial feature maps of task-specific Adapters for three NAS components. 
    Each row shows the input patch and the down-projection, normalized, and up-projection 
    feature maps of the corresponding Adapter.}
    \label{fig:adapter_spatial}
    \vspace{-8mm}
\end{figure}

To check whether this subspace-decoupling actually happens in practice, we further visualize task-specific Adapter activations in the last Swin-T stage (Sec.~\ref{sec:experiment}, Fig.~\ref{fig:adapter_spatial}). As shown later, different tasks attend to partially disjoint yet pathology-consistent regions, providing qualitative evidence that $L_{\text{ortho}}$ indeed drives the Adapters to exploit distinct feature subspaces.

\subsection{Uncertainty-Weighted Multi-Task Objective}

We optimize all tasks jointly with a multi-task loss that follows the homoscedastic uncertainty weighting scheme of Kendall et al.~\cite{kendall_mtl}. Concretely, for each task $t$ we keep a learnable noise variance $\sigma_t^2$ and combine the per-task losses $L_t$ (cross-entropy or focal loss) as
\begin{equation}
\label{eq:mtl}
L_{\text{MTL}} = \sum_{t=1}^{T} \left( \frac{1}{2\sigma_t^2} L_t + \frac{1}{2} \log \sigma_t^2 \right),
\end{equation}
where $T=3$ for steatosis, ballooning, and inflammation. During training, we directly optimize $\log \sigma_t^2$ (clamped to a fixed range for numerical stability), so that tasks with larger predictive noise automatically receive smaller weights via $1/(2\sigma_t^2)$, while the $\frac{1}{2}\log \sigma_t^2$ term regularizes the scale and prevents trivial solutions. This module is introduced because the three NAS components differ in label noise, class imbalance, and difficulty; manually tuning task weights is brittle. The uncertainty-based formulation lets the network learn task weights from data, balancing gradients across tasks and providing a principled complement to our representation-level orthogonal constraint.

\subsection{Training Procedure and Output Heads}

We jointly optimize all components with a single training objective that couples the uncertainty-weighted task losses and the orthogonal regularizer. Given per-task classification losses $L_t$ and the uncertainty-weighted aggregate $L_{\text{MTL}}$ in Eq.~\ref{eq:mtl}, the final objective is

\begin{equation}
L_{\text{total}} = L_{\text{MTL}} + \lambda L_{\text{ortho}},
\end{equation}
where $\lambda$ is a modest constant tuned on the validation set to keep the orthogonality term at the same order of magnitude as the task losses. In optimization, we freeze the Swin-T backbone and only update the task-specific Adapters / LoRA modules, the classification heads, and the uncertainty parameters $\sigma_t$; this realizes parameter-efficient fine-tuning while focusing capacity on task-dependent subspaces.

On top of the adapted shared features, each task uses an independent output head (LayerNorm → GELU-activated fully connected layer → dropout with $p=0.5$ → final classifier) to produce logits. This module is designed so that all tasks can reuse a common backbone representation, but their last-mile mappings remain disentangled: there is no explicit cross-task fusion at the head level, so interference is mainly controlled by $L_{\text{MTL}}$ (balancing gradient magnitudes) and $L_{\text{ortho}}$ (separating feature subspaces). In practice, this combination yields more stable training than manual task reweighting alone and empirically leads to better NAS sub-score prediction.

\section{Dataset}

\subsection{Animal Model and Ethical Compliance}
All histological data were obtained from a controlled mouse NAFLD model. Wild-type mice were used to induce NAFLD/NASH phenotypes through a 60\% high-fat diet following established protocols. All experimental procedures were approved by the relevant institutional ethics committee and conducted in accordance with applicable animal welfare regulations.

\subsection{Histopathology Acquisition and Inclusion Criteria}
H\&E-stained whole-slide images (WSIs) were digitized using a Hamamatsu scanner (Hamamatsu Photonics, Japan). Cropped regions were extracted into $416\times416$ RGB patches. To ensure annotation reliability and morphological consistency, we retained only patches that clearly exhibited NAFLD-related histological features (e.g., steatosis or inflammatory infiltrates), had complete and unambiguous scoring information for all NAS components, and showed consistent staining quality without visible artifacts. Patches that did not meet these criteria were excluded from further analysis.

To enhance the robustness of model training, we further expanded the dataset through standard rotation and flip augmentations, resulting in an overall 8-fold expansion \cite{perez2017effectiveness}, as shown in Fig.~\ref{fig:architecture}. In total, the curated dataset comprises 3,192 qualified image patches after screening.

\subsection{Scoring Protocol and Quality Control}
Each patch was evaluated following the standard NAS scoring system, as summarized in Table~\ref{tab:nas_core}. All annotations were reviewed by domain experts, with discrepancies resolved through consensus.

\begin{table}[t]{
  \centering
  \caption{NAS Scoring Criteria (Core Histological Features)}
  \vspace{-7mm}
  \label{tab:nas_core}
  \resizebox{\linewidth}{!}{
    \begin{tabular}{l c c}
      \toprule
      \textbf{Histological Feature} & \textbf{Extent} & \textbf{Score} \\
      \midrule
      \multirow{4}{*}{Steatosis (\%)} 
        & $<5$        & 0 \\
        & $5$--$33$   & 1 \\
        & $34$--$66$  & 2 \\
        & $>66$       & 3 \\
      \midrule
      \multirow{3}{*}{Hepatocellular Ballooning} 
        & None                & 0 \\
        & Few ballooned cells & 1 \\
        & Numerous ballooned cells & 2 \\
      \midrule
      \multirow{4}{*}{Lobular Inflammation} 
        & None      & 0 \\
        & $<2$ foci & 1 \\
        & $2$--$4$ foci & 2 \\
        & $>4$ foci & 3 \\
      \bottomrule
    \end{tabular}
  }
}
\end{table}

\subsection{Statistical Characteristics}
\begin{figure}[t]
    \centering
    \includegraphics[width=1.0\linewidth]{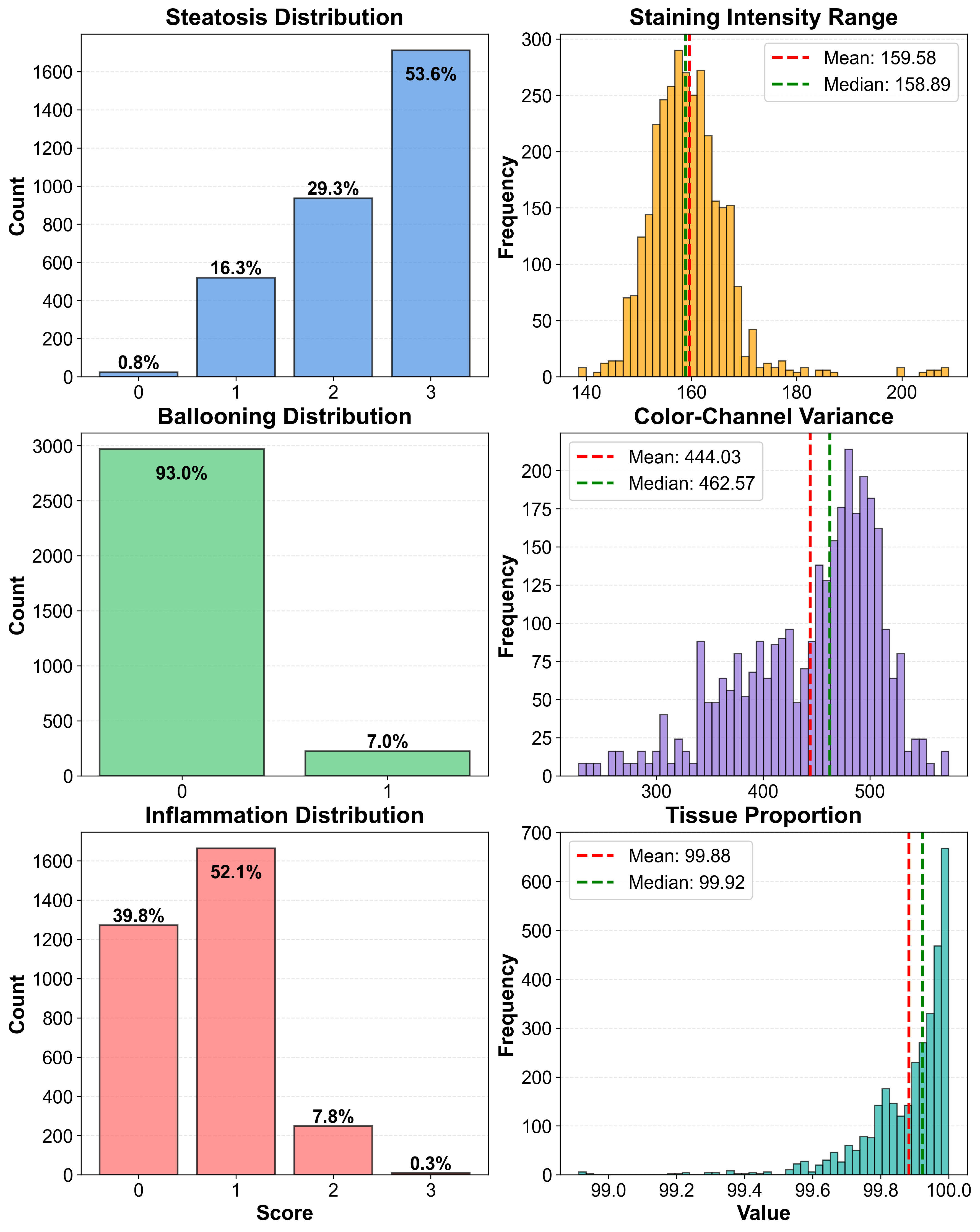}
    \vspace{-3mm}
    \caption{Summary of NAS class distributions and patch-level quality statistics in our dataset.}
    \label{fig:dataset_stats}
    \vspace{-5mm}
\end{figure}

Fig.~\ref{fig:dataset_stats} summarizes the class distributions of NAS indicators and key patch-level quality statistics. The dataset exhibits clear class imbalance across NAS components while maintaining consistently high tissue coverage, highlighting both the learning challenges and data quality.

\section{Experiment}
\label{sec:experiment}

\subsection{Implementation Details}
For a fair comparison, all methods use the same training/validation split and optimization pipeline. Experiments are conducted on NVIDIA RTX 3090-ti GPU. All liver histology patches are resized from $416 \times 416$ to $224 \times 224$ and normalized with ImageNet statistics. We only apply random horizontal and vertical flips as data augmentation. The batch size is fixed to 16.

We train all models with AdamW (initial learning rate $8 \times 10^{-5}$, weight decay $0.01$) for 200 epochs. The learning rate is reduced using a plateau-based schedule with a multiplicative decay factor of $0.3$ when the validation metric stops improving. To reduce the impact of randomness, we fix a global random seed and repeat each configuration three times, reporting the average performance across runs.

All multi-task variants are optimized with the loss defined in Sec.~\ref{sec:method}: a per-task classification loss combined via homoscedastic uncertainty weighting and augmented with an orthogonal subspace regularizer. In the main experiments, we set the task weights for steatosis, ballooning, and inflammation to $(1.0, 1.2, 1.1)$ to mildly up-weight the rarer and more difficult tasks while keeping the overall loss scale comparable across tasks.

\subsection{Evaluation Results}

\subsubsection{Backbone Comparison}

We compare three pretrained visual encoders for the multi-task NAS scoring problem, namely Swin-Tiny-22k, Swin-Small-22k, and DINOv2 ViT-Small~\cite{Liu_2021_ICCV,oquabDINOv2LearningRobust2024}. All models share the same multi-task heads and loss configuration; only the shared backbone is replaced.

Table~\ref{tab:backbone_combined} reports per-task accuracy on our NAFLD dataset. Swin-Tiny achieves the best performance on \emph{steatosis} and \emph{inflammation}, while \emph{ballooning} is relatively easy and almost saturated for all backbones.

\begin{table}[t]
\centering
\caption{Backbone comparison on our NAFLD dataset and the Farzi NAFLD dataset.}
\label{tab:backbone_combined}
\setlength{\tabcolsep}{4pt}
\renewcommand{\arraystretch}{1.05}
\begin{tabular}{lccc}
\toprule
\multicolumn{4}{c}{\textbf{Ours NAFLD dataset}} \\
\midrule
Backbone   & Steatosis Acc & Ballooning Acc & Inflammation Acc \\
\midrule
Swin-Tiny  & 0.9050        & 0.9625         & 0.8262 \\
Swin-Small & 0.8460        & 0.9250         & 0.7420 \\
DINOv2     & 0.7650        & 0.9526         & 0.7250 \\
\midrule
\multicolumn{4}{c}{\textbf{Farzi NAFLD dataset}} \\
\midrule
Backbone   & Steatosis Acc & Ballooning Acc & Inflammation Acc \\
\midrule
Swin-Tiny  & 0.8080        & 0.9108         & 0.7226 \\
Swin-Small & 0.7087        & 0.9139         & 0.6783 \\
DINOv2     & 0.7907        & 0.9146         & 0.7501 \\
\bottomrule
\end{tabular}
\vspace{-5mm}
\end{table}

To assess transferability, we repeat the backbone comparison on the public NAFLD dataset of Farzi \emph{et al.}~\cite{heinemann2019deep}, using the same multi-task heads and orthogonal decoupling configuration resulting in Table~\ref{tab:backbone_combined}. On this dataset, DINOv2 attains slightly higher accuracy on \emph{ballooning} and \emph{inflammation}, while Swin-Tiny remains competitive. Overall, these results indicate that our subspace-decoupled multi-task framework is robust to the choice of backbone, and we adopt Swin-Tiny as the default due to its favorable trade-off between performance and model size.

\subsubsection{Comparisons with Other Methods}

We then compare the proposed method with Farzi \emph{et al.}'s InceptionV3-based CNN, where separate classification networks are trained independently for \emph{steatosis}, \emph{ballooning}, and \emph{inflammation} ~\cite{heinemann2019deep}. 

 As shown in Table~\ref{tab:reference_comparison}, our method achieves comparable than the InceptionV3-based CNN on  \emph{steatosis} and \emph{ballooning}. For the more difficult and label-noisy \emph{inflammation} task, our approach attains a similar level of performance while offering the advantages of a unified multi-task formulation. 
 
 

\begin{table}[t]
    \centering
    \caption{Comparison between our method and the InceptionV3 on the NAFLD dataset.}
    \label{tab:reference_comparison}
    \begin{tabular}{lccc}
        \toprule
        Method                               & Steatosis Acc & Ballooning Acc & Inflammation Acc \\
        \midrule
        InceptionV3 & 0.8722        & 0.9061        & 0.8027 \\
        Ours      & \textbf{0.9050} & \textbf{0.9625} & \textbf{0.8262} \\
        \bottomrule
    \end{tabular}
    \vspace{-5mm}
\end{table}

\subsection{Ablation Studies}
\subsubsection{Orthogonal Constraint Weight $\lambda$}
We specifically study the effect of the orthogonal regularization to isolate its role in mitigating negative transfer among correlated NAS tasks. We fix the backbone and Adapter configuration and vary the orthogonal regularization weight $\lambda \in \{0, 0.01, 0.1, 1.0\}$. The results are reported in Table~\ref{tab:ablation_combined}.

\begin{table}[t]
\centering
\caption{Ablation studies on orthogonal regularization weight $\lambda$ and structure design.}
\label{tab:ablation_combined}
\setlength{\tabcolsep}{4pt}
\renewcommand{\arraystretch}{1.05}
\begin{tabular}{lccc}
\toprule
\multicolumn{4}{c}{\textbf{Orthogonal regularization weight $\lambda$}} \\
\midrule
$\lambda$ & Steatosis Acc & Ballooning Acc & Inflammation Acc \\
\midrule
0.0  & 0.8327 & 0.9220 & 0.6720 \\
0.01 & 0.8632 & 0.9423 & 0.7460 \\
0.1  & \textbf{0.9050} & 0.9625 & 0.8262 \\
1.0  & 0.8525 & 0.9324 & 0.4500 \\
\midrule
\multicolumn{4}{c}{\textbf{Structure type}} \\
\midrule
Structure Type & Steatosis Acc & Ballooning Acc & Inflammation Acc \\
\midrule
Adapter-only & 0.9050 & 0.9625 & 0.8262 \\
LoRA-only    & 0.7875 & 0.9625 & \textbf{0.7624} \\
\bottomrule
\end{tabular}
    \vspace{-5mm}
\end{table}

Introducing a small orthogonal penalty ($\lambda>0$) consistently improves the steatosis accuracy while keeping the other two tasks unchanged, and $\lambda=0.1$ yields the best overall performance. A much larger weight ($\lambda=1.0$) does not bring further gains, suggesting that a moderate level of decoupling is sufficient to mitigate negative transfer without over-penalizing shared representations.

\subsubsection{Adapter versus LoRA}

We next study an alternative parameter-efficient strategy based on LoRA. Instead of inserting bottleneck Adapters after the MLP blocks, LoRA injects low-rank trainable updates into the key linear projections of each Transformer block, while keeping the Swin-T backbone weights frozen. For each NAS task, we maintain its own set of low-rank LoRA matrices, allowing attention weights to be adjusted in a task-specific manner with only a small number of additional parameters.

Table~\ref{tab:ablation_combined} shows that Adapter-only and LoRA-only achieve comparable performance, with LoRA slightly better on inflammation, so we use Adapter as the default and regard LoRA as a practical alternative.

\subsection{Qualitative Analysis of Task-Specific Adapter Activations}

To better understand what the task-specific Adapters learn, we visualize their spatial activations in the last Swin-T stage. For each task $t \in \{\text{steatosis}, \text{ballooning}, \text{inflammation}\}$, we tap into the Adapter at three internal points (down-projection, normalization, and up-projection). Given the final-stage feature grid of size $7\times 7\times d$, we compute the $\ell_2$ norm over the channel dimension at each spatial location, min–max normalize the resulting $7\times 7$ map to $[0,1]$, and upsample it to the input resolution to obtain a pseudo–color heatmap.

Fig.~\ref{fig:adapter_spatial} shows representative examples for the three NAS components. Each row corresponds to one task: the leftmost image is the input H\&E patch, followed by the activation maps at the three Adapter points. For steatosis, the responses concentrate around regions densely populated with lipid vacuoles, while the background tissue is largely suppressed. For ballooning, activations shift towards swollen hepatocytes and the surrounding parenchyma. For inflammation, high-response areas align with clusters of inflammatory cells and peri-lobular structures. These partially disjoint yet pathology-consistent patterns indicate that different tasks indeed rely on distinct regions of the shared feature map, providing qualitative evidence that our task-specific, subspace-decoupled Adapters work as intended and complement the quantitative gains brought by the orthogonal regularizer.

\section{Conclusion}

In this work, we presented a parameter-efficient Subspace-Decoupling Vision Transformer framework for multi-task histological scoring of NAFLD/NASH. By integrating lightweight task-specific Adapters with an orthogonality-based regularization, the proposed method effectively mitigates negative transfer among correlated NAS components while preserving shared low-level representations in a frozen ViT backbone. Extensive experiments demonstrate that this design achieves improved and more balanced performance across tasks with only a minimal increase in trainable parameters. In addition, we will release a curated multi-task mouse NAFLD histology dataset with detailed annotations to support reproducible research in automated liver pathology.

\textbf{Limitations and Future Work.}
This work focuses on patch-level NAS component prediction and does not explicitly address slide-level aggregation or patient-level diagnosis, which are important for clinical translation. In addition, experiments are conducted primarily on a mouse NAFLD model; further validation on large-scale, multi-center human cohorts is required to assess generalizability. Finally, extending the framework to incorporate additional pathological or clinical endpoints beyond NAS represents a promising direction for future research.
 
\section{Acknowledgments}

We sincerely thank Yuan Liang and Yuting Liu (Department of Pathology, School of Basic Medical Sciences, Capital Medical University) for their guidance and support throughout this study. We also thank Maoqi Liu, a graduate student at Beijing University of Posts and Telecommunications, for helpful comments on the manuscript.

\bibliographystyle{IEEEbib}
\bibliography{icme2026references}

\end{document}